\documentclass{article}

     \usepackage[final,nonatbib]{neurips_2019}

\usepackage[utf8]{inputenc} 
\usepackage[T1]{fontenc}    
\usepackage{hyperref}       
\usepackage{graphicx}
\usepackage{url}            
\usepackage{booktabs}       
\usepackage{amsfonts}       
\usepackage{nicefrac}       
\usepackage{microtype}      
\usepackage{subcaption}
\usepackage{caption}
\usepackage{siunitx}
\usepackage{authblk}
\graphicspath{ {./images/} }

\title{Representing Ambiguity in Registration Problems \\with Conditional Invertible Neural Networks}

\author[1]{Darya Trofimova} 
\author[1,2]{Tim Adler}
\author[3]{Lisa Kausch}
\author[4]{Lynton Ardizzone}
\author[3]{Klaus Maier-Hein}
\author[4]{Ulrich Köthe}
\author[4]{Carsten Rother}
\author[1,2,5]{Lena Maier-Hein}
\affil[1]{Division of Computer Assisted Medical Interventions \protect\\German~Cancer~Research~Center, Heidelberg, Germany \protect\\(\texttt{d.trofimova|t.adler|l.maier-hein@dkfz.de})}
\affil[2]{Faculty of Mathematics and Computer Science, Heidelberg~University,~Heidelberg,~Germany}
\affil[3]{Division of Medical Image Computing, German~Cancer~Research~Center, Heidelberg, Germany}
\affil[4]{Visual Learning Lab, Heidelberg University,~Heidelberg,~Germany}
\affil[5]{Medical Faculty, Heidelberg University, Heidelberg, Germany}

\begin{document}
\maketitle
\section{Introduction}
Image registration is the basis for many applications in the fields of medical image computing and computer assisted interventions. One example is the registration of 2D X-ray images with preoperative three-dimensional computed tomography (CT) images in intraoperative surgical guidance systems. Due to the high safety requirements in medical applications, estimating registration uncertainty is of a crucial importance in such a scenario. However, previously proposed methods, including classical iterative registration methods~\cite{registration_iterative, 563664, Viola1997} and deep learning-based methods~\cite{registration_dl, salehi2018, miao2016, yang2016} have one characteristic in common: They lack the capacity to represent the fact that a registration problem may be inherently ambiguous, meaning that multiple (substantially different) plausible solutions exist. To tackle this limitation, we explore the application of invertible neural networks (INN) as core component of a registration methodology. In the proposed framework, INNs enable going beyond point estimates as network output by representing the possible solutions to a registration problem by a probability distribution that encodes different plausible solutions via multiple modes. In a first feasibility study, we test the approach for a 2D/3D registration setting by registering spinal CT volumes to X-ray images. To this end, we simulate the X-ray images taken by a C-Arm with multiple orientations using the principle of digitially reconstructed radiographs (DRRs)~\cite{unberath2018deepdrr}. Due to the symmetry of human spine, there are potentially multiple substantially different poses of the C-Arm that can lead to similar projections. The hypothesis of this work is that the proposed approach is able to identify multiple solutions in such ambiguous registration problems.

\begin{figure}[!htbp]
  \centering
  \subfloat[Schematic overview of the approach]{\includegraphics[width=0.51\textwidth]{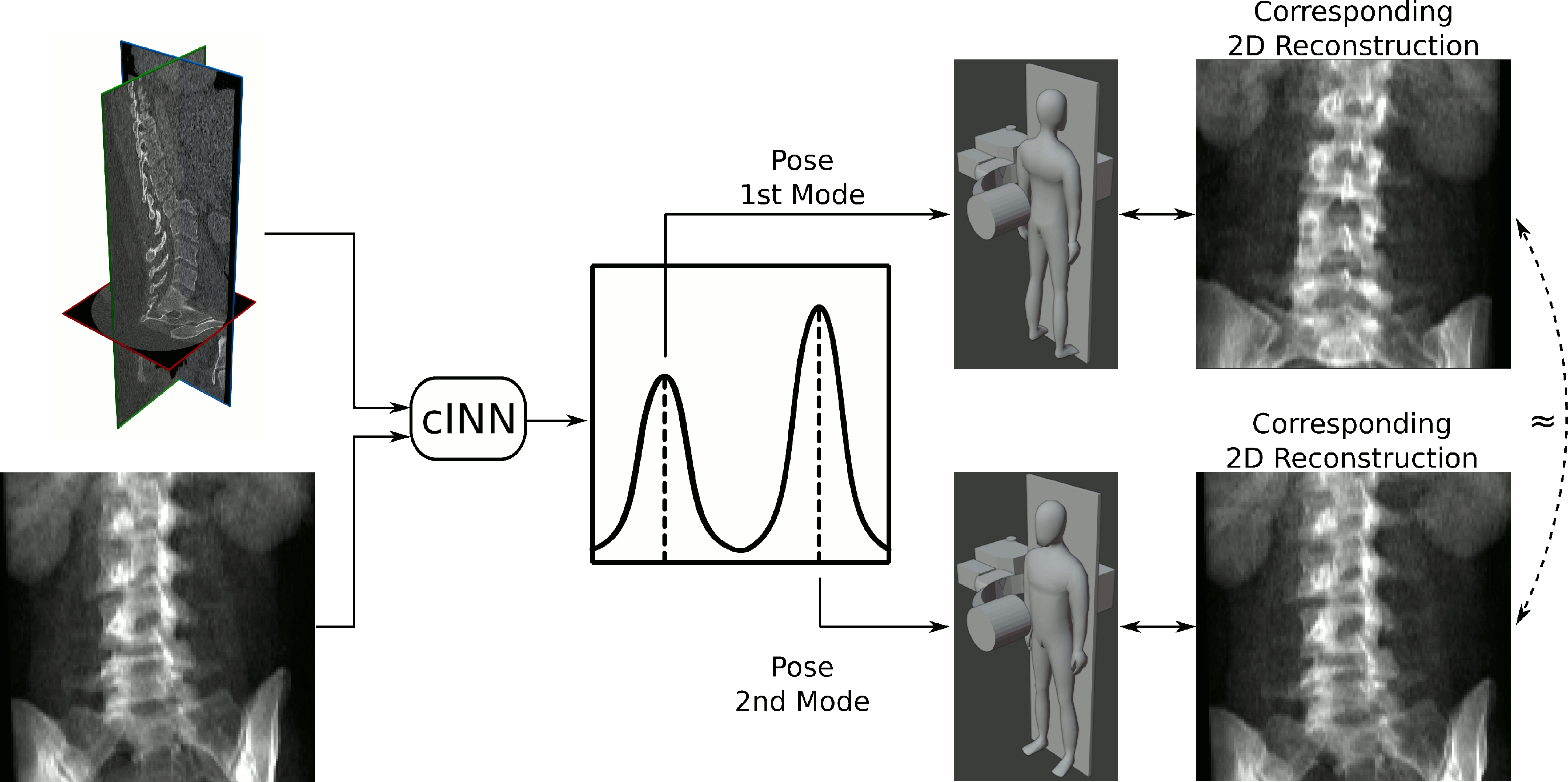}\label{schema_setting}}
  \hfill
  \subfloat[Model architecture]{\includegraphics[width=0.48\textwidth]{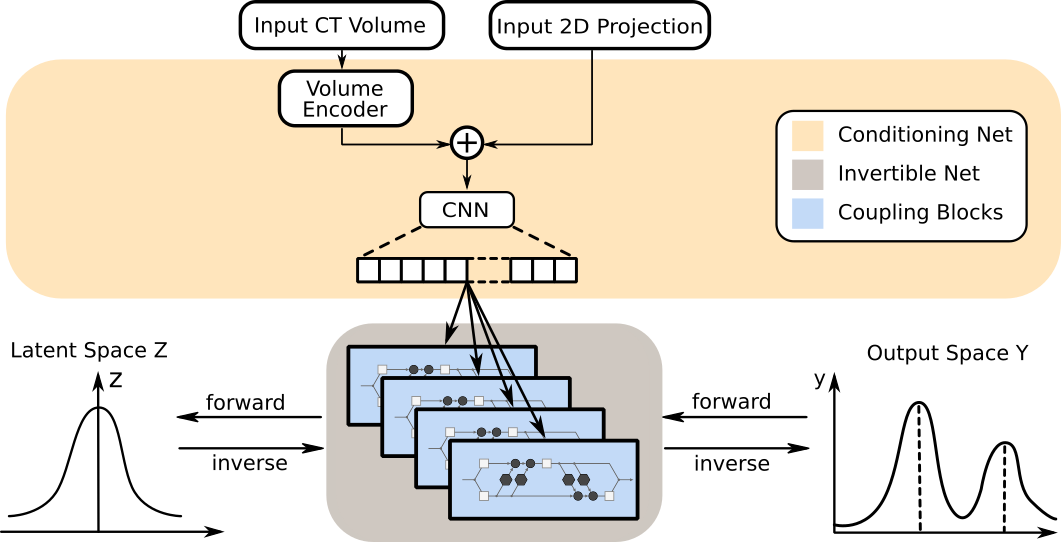}\label{fig:f2}}
  \caption{2D/3D registration approach using a conditional invertible neural network (cINN) as core component to represent ambiguity. (a) shows a schematic illustration of the approach where  two different poses of the patient lead to similar projections and associated plausible modes identified by the model. (b) describes the flow of the data through the main components of the model.}
\end{figure}

\section{Methodology}

INN architectures, such as~\cite{ardizzone_analyzing_2018} have recently been proposed for the analysis of potentially ill-posed inverse problems. In contrast to common neural network architectures, they output representations of the full probability density function rather than point estimates representing only a single solution. As the originally proposed architecture for INNs is not easily transferable to large input sizes and suffers from instabilities resulting from zero-padding, we base our work on a more recent variant, referred to as conditional INNs (cINNs)~\cite{ardizzone_conditional}. This architecture combines an INN with an unconstrained feed-forward network for conditioning (\textit{conditioning network}) and features a stable, maximum likelihood-based training procedure for jointly optimizing the parameters of the INN and the conditioning network. A schematic overview of the model architecture applied to our setting can be found in Figure~\ref{fig:f2}.  Following~\cite{kausch2020, c_arm_coords}, we represent a rigid transformation by three translation and two rotation parameters. The output of the cINN is a probability density function for these parameters (see Fig.~\ref{schema_setting}). CT volume and 2D projection are used as conditioning input and are jointly projected to a compact representation by a standard feed forward convolutional neural network (CNN). 
This representation, along with samples drawn from the latent space Z, is received by the coupling blocks of the INN which outputs a conditional probability distribution of the pose parameters.   

The following paragraphs explain the main building blocks of our architecture and performed training stages.

\paragraph{Conditioning Network}
To eliminate the need for the coupling blocks to learn the complete representation of the input images, a conditioning network is applied that transforms the two input images to an intermediate representation. The choice of the architecture of the conditioning network was inspired by~\cite{airnet}, where core elements of the registration network are blocks with convolutional layers followed by batch normalization, dropout layers and rectified linear unit (ReLU) activations. In the first stage of the training, we pre-train the conditioning network with MSE loss to predict the pose parameters.

\paragraph{Conditional Invertable Neural Network}
 We base our cINN architecture on~\cite{ardizzone_conditional} and the implementation on the corresponding PyTorch package\footnote{\url{https://github.com/VLL-HD/FrEIA}}. In this first feasibility study, training is performed with a maximum likelihood loss (enforcing a standard Gaussian in the latent space), batch size of 32, learning rate of 0.01, step decay of the learning rate every 100 epochs, and the Adam optimizer with weight decay. During this second training stage, the conditioning network is further optimized together with the cINN. We include noise and contrast augmentation for both CT volume and 2D projections. In addition, we use soft clamping of the scale coefficients within the coupling blocks to restrain the model to stabilize training.
Upon test time, CT volume and 2D projection serve as conditioning input, and repeated sampling from the latent space results in a full posterior over the parameter space. 

\section{Experiments and Results}
For validation of our ambiguity-aware registration framework, we picked a clinical use case in which we expected multiple plausible registration solutions, namely the registration of 2D spine C-arm images with 3D CT volume data. In this setting, ambiguity results from the general symmetry of the spine. The purpose of our experiments was  to demonstrate the capability of the method to detect multiple plausible registration solutions.

\paragraph{Dataset}
In this first feasibility study, we used the UWSpine dataset~\cite{dataset1, dataset2} which comprises spine-focused CT volumes of 125 patients.  We transformed the volumes to a homogeneous voxel spacing and discarded those images smaller than 128x256x128. For every CT volume, we sampled 100 different poses  of the C-Arm device and computed corresponding DRRs. The parameters representing the virtual C-Arm pose were determined as follows: The translation along the sagittal, longitudinal and transverse axis was randomly sampled from a continuous uniform distribution with range [-20 mm, 20 mm].  The two angles representing the rotation around the longitudinal (LAO) and transverse (CRAN) axis of the patient were sampled from a discrete uniform distribution with range [-\ang{20}, \ang{20}] with a step of \ang{1} (which is similar to the rotation ranges of typical C-Arm machine). In addition, with even odds the LAO angle was shifted by \ang{180} to introduce a possible ambiguity in the projections (see Fig.~\ref{schema_setting}).

\paragraph{cINN-based registration and mode detection} 
We applied the framework introduced in the previous section for registration of the DRRs with the 3D volume. Owing to our simulation setting we expected either one main mode (in the case of an asymmetric spine/setup) or two main modes (in the case of a symmetric spine/setup) in the registration solution. To quantify the number of modes, we fitted the estimated posteriors to Gaussian Mixture Models (GMMs) with a single component GMM(n=1) and with two components GMM(n=2). By comparing the Akaike information criterion (AIC) for both models, we labeled the sample either as \emph{multi-modal}, if $\text{AIC}(n=2) < \text{AIC}(n=1) - 2000$, or as \emph{uni-modal}, otherwise. We then determined the registration parameters for each of the modes by estimating the means of the Gaussian distributions. While this is a pragmatic way of extracting the modes from the network data, further research is required to optimize this step in the pipeline with respect to robustness.

\paragraph{Validation} 
Due to the lack of ground truth information on the number and characteristics of plausible solutions, we decided to base our validation on the reprojection error as proxy for registration performance. More specifically, out of $N=2200$ samples in our test set we identified $n=686$ cases in which multiple modes were detected. We then re-simulated DRRs with the device poses corresponding to the different modes and computed the L1 norm between the re-simulated projection and the ground truth projection. As a comparison, we also fitted a single Gaussian (representing a single valid solution) to the network output and computed the L1 distance for the corresponding reprojection to the ground truth. The reprojection error for the multiple modes was $\mathnormal{0.104}$ on average ($\mathnormal{0.082}$ for the mode closer to the ground truth parameters and $\mathnormal{0.125}$ for the second one) compared to $\mathnormal{0.128}$ obtained with a single Gaussian. 

\begin{figure}
\centering
\includegraphics[width=1\linewidth]{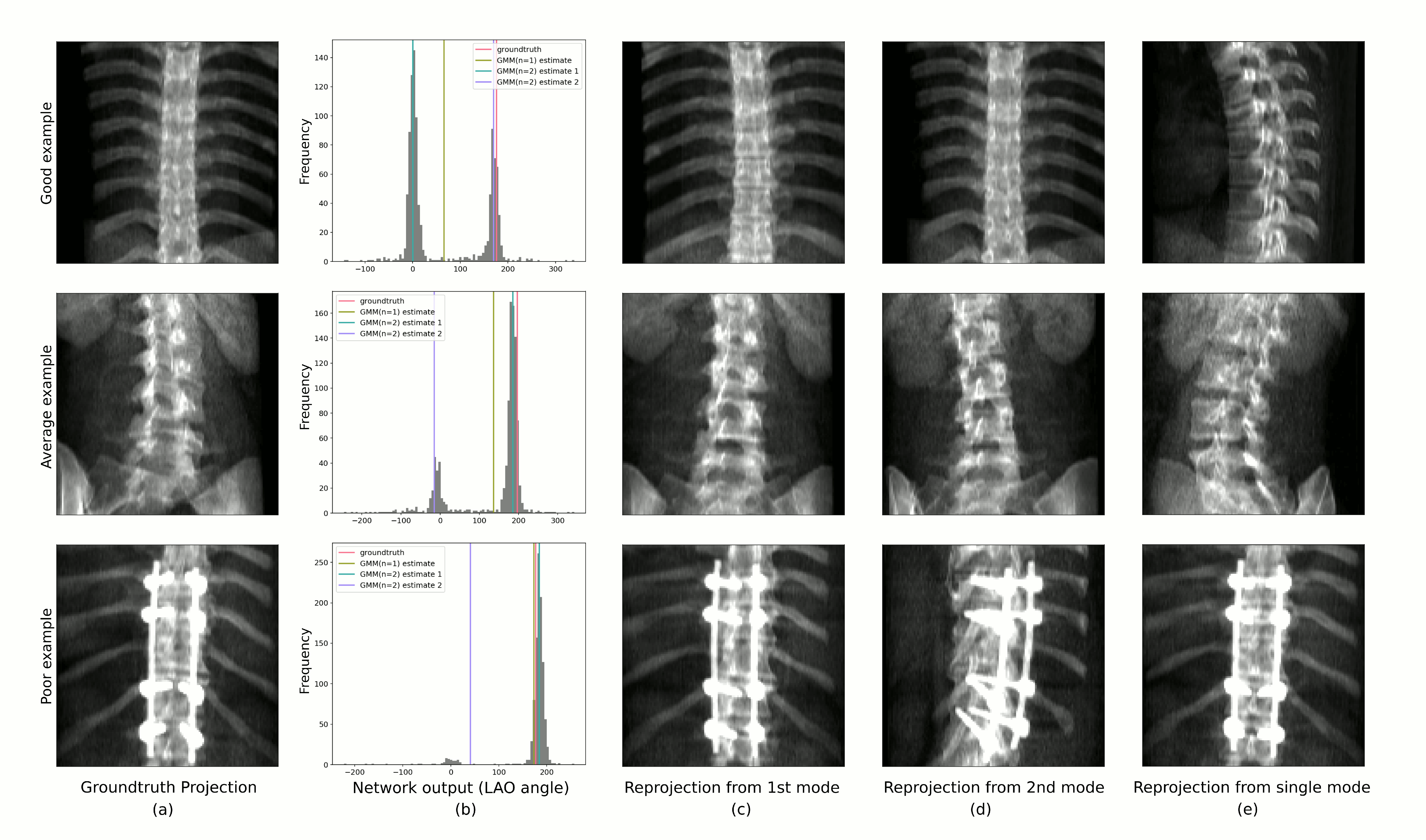}
\caption{Examples of registration results. From left to right: (a) Ground truth projections, (b) posterior of LAO angle, (c-d) reprojections corresponding to the first and second mode and (e) reprojection corresponding to the main mode when only one Gaussian is fitted to the data. The different rows correspond to a good (top), an average (middle) and a poor (bottom) example.}
\label{resimulated_images}
\end{figure}

In Fig.~\ref{resimulated_images} (top) a good example of two strongly visible modes is shown. The centers of the calculated posteriors are estimated correctly with GMM(n=2) leading to similar looking reprojections even with significantly distinct poses. The  GMM(n=1) estimate falls in a zone of very low probability leading to a bad pose estimation and a bad reprojection. A similar scenario with a multi-modal posterior distribution featuring one strong primary and one secondary mode is show in Fig.~\ref{resimulated_images} (middle). The reprojections of both modes are close to the ground truth whereas the GMM(n=1) estimate is off. Fig.~\ref{resimulated_images} (bottom) shows a failure case of our method. The second mode is very small and most likely can be regarded a false positive with respect to the mode detection. Furthermore, the corresponding node center is not correctly estimated leading to a poor reprojection. The dominant mode and the GMM(n=1) estimate are both close to the ground truth and lead to good reprojections.

\paragraph{Interpretation} 
Our experiments indicate that cINNs are a possible solution for addressing ambiguity in registration problems. Further work on evaluation metrics as well as testing the approach in a real world setting needs to be invested.

\section{Broader Impact Statement}
Uncertainty handling in deep learning-based image analysis is gaining increasing attention and may be a crucial factor for clinical translation of research. So far, a lot of uncertainty research has been focusing on epistemic uncertainty arising from insufficient training data as well as aleatoric uncertainty, defined as the potential intrinsic randomness of the data generation process. A third type of uncertainty that has received very little attention in the literature is the potential inherent ambiguity of the problem. State-of-the-art approaches to image interpretation typically provide point estimates and neglect the fact that the problem may be ill-posed. Consequently, the estimations cannot generally be trusted to be close to the ground truth. This work is, to our knowledge, the first to address this problem in the specific context of intraoperative medical image registration. Based on the principle of invertible neural networks we present a framework for representing multiple plausible solutions via multiple modes in the output data. The work could become an important first step in handling ambiguities in registration problems.

\subsubsection*{Acknowledgments}

This project has received funding from the European Union Horizon 2020 research and innovation program through the European Research Council (ERC) starting grant COMBIOSCOPY under Grant Agreement No. ERC-2015-StG-37960. It was also supported from the European Union Horizon 2020 research and innovation program through the ERC grant under Grant Agreement No. 647769 as well as from the Federal Ministry of Education and Research of Germany project High Performance Deep Learning Framework (No 01IH17002).

\medskip
\small
\bibliography{dtrofimova_neurips2020.bib}
\end{document}